%% file: paper.tex
\documentclass{article}

\usepackage{microtype}
\usepackage{graphicx}
\usepackage{subfigure}
\usepackage{booktabs} 
\usepackage{dsfont}
\usepackage{amsfonts}
\usepackage{mathtools}
\usepackage[separate-uncertainty=true,table-align-uncertainty=true]{siunitx}
\usepackage{lipsum}

\usepackage{hyperref}

\newcommand{\ones}[1]{\mathbf{1}_{#1}}
\newcommand{\zeros}[1]{\mathbf{0}_{#1}}

\usepackage[accepted]{icml2021}

\definecolor{gray}{rgb}{0.5,0.5,0.5}

\newcommand{\minihead}[1]{{\noindent\textbf{#1.} }}

\renewcommand{\cite}[1]{\citep{#1}}

\icmltitlerunning{Sinkhorn Label Allocation: Semi-Supervised Classification via Annealed Self-Training }
\begin{document}

\twocolumn[
\icmltitle{Sinkhorn Label Allocation: \\Semi-Supervised Classification via Annealed Self-Training }



\icmlsetsymbol{equal}{*}

\begin{icmlauthorlist}
\icmlauthor{Kai Sheng Tai}{stanford}
\icmlauthor{Peter Bailis}{stanford}
\icmlauthor{Gregory Valiant}{stanford}
\end{icmlauthorlist}

\icmlaffiliation{stanford}{Stanford University, Stanford, CA, USA}

\icmlcorrespondingauthor{Kai Sheng Tai}{kst@cs.stanford.edu}


\vskip 0.3in
]



\printAffiliationsAndNotice{}  

\input{abstract}
\input{introduction}

\input{method}

\input{related}

\input{experiments}
\input{discussion}

\section*{Acknowledgements}

This research was supported in part by affiliate members and other supporters of the Stanford DAWN project---Ant Financial, Facebook, Google, and VMware---as well as Toyota Research Institute, Northrop Grumman, Cisco, and SAP. This work was also supported by NSF awards 1704417 and 1813049, ONR YIP award N00014-18-1-2295, and DOE award DE-SC0019205. Any opinions, findings, and conclusions or recommendations expressed in this material are those of the authors and do not necessarily reflect the views of the National Science Foundation. Toyota Research Institute (``TRI'') provided funds to assist the authors with their research but this article solely reflects the opinions and conclusions of its authors and not TRI or any other Toyota entity.

\bibliography{references}
\bibliographystyle{icml2021}

\clearpage
\input{appendix}

\end{document}

%% file: abstract.tex
\begin{abstract}
Self-training is a standard approach to semi-supervised learning where the learner's own predictions on unlabeled data are used as supervision during training.
In this paper, we reinterpret this label assignment process as an optimal transportation problem between examples and classes, wherein the cost of assigning an example to a class is mediated by the current predictions of the classifier.
This formulation facilitates a practical annealing strategy for label assignment and allows for the inclusion of prior knowledge on class proportions via flexible upper bound constraints.
The solutions to these assignment problems can be efficiently approximated using Sinkhorn iteration, thus enabling their use in the inner loop of standard stochastic optimization algorithms.
We demonstrate the effectiveness of our algorithm on the CIFAR-10, CIFAR-100, and SVHN datasets in comparison with FixMatch, a state-of-the-art self-training algorithm.
\end{abstract}

%% file: introduction.tex
\section{Introduction}

In semi-supervised learning (SSL), we are given a partially-labeled training set consisting of labeled examples $\{(x_i, y_i) \mid i = 1, \dots, n_\ell\}$ and unlabeled examples $\{ x_i \mid i = n_{\ell} + 1, \dots, n \}$, with $x \in \mathcal{X}$ and $y\in\mathcal{Y}$.
Our goal in this setting is to leverage our access to unlabeled data in order to learn a predictor $f: \mathcal{X} \rightarrow \mathcal{Y}$ that is more accurate than a predictor trained using the labeled data alone.
This setup is motivated by the high cost of obtaining human annotations in practice, which results in a relative scarcity of labeled examples in comparison with the total volume of unlabeled data available for training.
Consequently, we are typically interested in the regime where $n_\ell \ll n$.

This paper focuses on \emph{self-training} for semi-supervised classification tasks.
Self-training, also known as self-labeling, is an SSL method where the classifier's own predictions on unlabeled data are used as additional supervision during training.
Specifically, self-training involves the following alternating process: in each iteration, the classifier's outputs are used to assign labels to unlabeled examples; 
these artificially labeled examples are then used as supervision to update the parameters of the classifier.
This intuitive bootstrapping procedure was first studied in the signal processing and statistics communities~\citep{scudder1965probability,mclachlan1975iterative,widrow1977stationary,nowlan1993soft} and was later adopted for natural language processing~\citep{yarowsky1995unsupervised,blum1998combining,riloff2003learning} and computer vision applications~\citep{rosenberg2005semi}.
More recently, methods based on self-training have been used to achieve strong empirical results on semi-supervised image classification tasks~\citep{xie2020self,fixmatch}.

The label assignment step is critical to the success of self-training. 
Incorrect assignments during training may cause further misclassifications in subsequent iterations, resulting in a feedback loop of self-reinforcing errors that ultimately yields a low-accuracy classifier.
As a result, self-training algorithms commonly incorporate various heuristics for mitigating label noise.
For instance, the state-of-the-art FixMatch algorithm~\citep{fixmatch} uses a \emph{confidence thresholding} rule wherein gradient updates only involve examples that are classified with a model probability above a user-defined threshold. 

\begin{figure*}[t]
\begin{center}
    \includegraphics[width=0.95\textwidth]{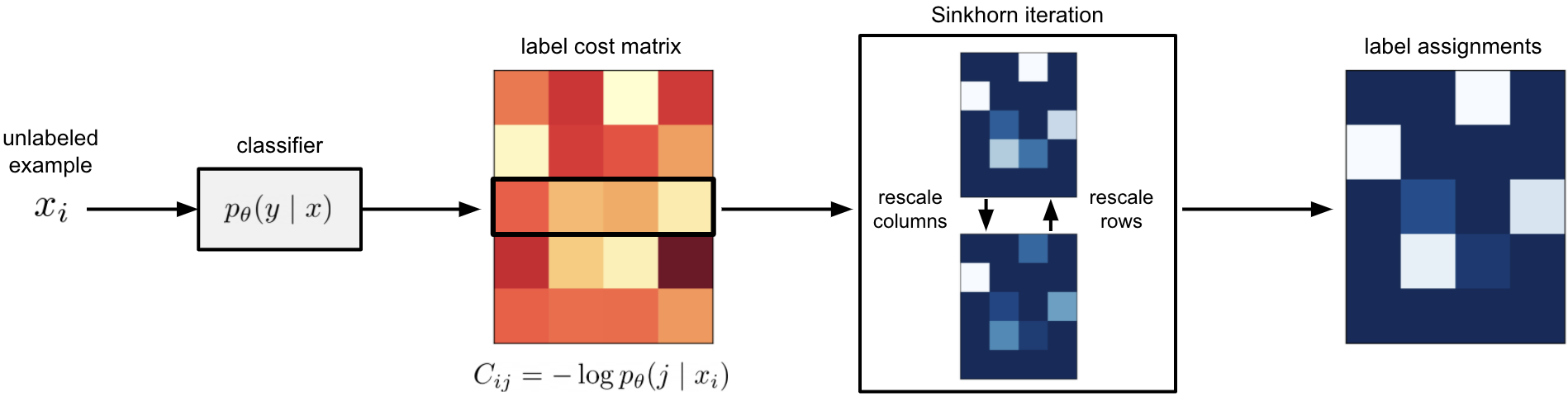}
\end{center}
\vspace{-1em}
\caption{\textbf{A schematic representation of the label assignment process in Sinkhorn Label Allocation (SLA).}
We model the label assignment task as an optimal transport problem between $n$ examples and $k$ classes, where the entries of the $n \times k$ assignment cost matrix are determined by the predictions of the classifier on unlabeled examples.
In the figure, lighter shades correspond to lower costs and higher label assignment weights.
By approximating the solution to the optimization problem using Sinkhorn iteration, we derive soft labels that can be used within a self-training algorithm.
SLA allows for additional control over the label assignment process through the use of constraints on class proportions and on the total mass of allocated labels.}
\label{fig:schematic}
\end{figure*}

Our main contribution is a new label assignment method, Sinkhorn Label Allocation (SLA), that models the task of matching unlabeled examples to labels as a convex optimization problem.
More precisely: in a classification problem where $\mathcal{Y} = \{1, \dots, k \}$, we seek an assignment $Q\in\mathbb{R}^{n\times k}$ of $n$ examples to $k$ classes that minimizes the total assignment cost $\sum_{ij} Q_{ij} C_{ij}(\theta)$, where the cost $C_{ij}(\theta)$ of assigning example $i$ to class $j$ is given by the corresponding negative log probability under the model distribution~$p_\theta$:
\begin{equation}
\label{eq:cost-matrix}
    C_{ij}(\theta) = - \log p_\theta ( j \mid x_i ).
\end{equation}
This formulation is desirable for several reasons.
First, we are able to subsume several commonly used label assignment heuristics within a single, principled optimization framework through our choice of constraints on the label assignment matrix $Q$. 
In addition to the aforementioned confidence thresholding heuristic, SLA is also able to simulate \emph{label annealing} strategies where the labeled set is slowly grown over time~(e.g., \citet{blum1998combining}), as well as \emph{class balancing} heuristics that constrain the artificial label distribution to be similar to the empirical distribution of the labeled set~(e.g., \citet{joachims1999transductive,kurakin2020remixmatch,xie2020self}).
Second, we can efficiently find approximate solutions for the resulting family of optimization problems using the Sinkhorn-Knopp algorithm~\citep{cuturi2013sinkhorn}.
Consequently, we are able to run SLA within the inner loop of standard stochastic optimization algorithms while incurring only a small computational overhead.

We demonstrate the practical utility of SLA through an evaluation on standard semi-supervised image classification benchmarks.
On CIFAR-10 with 4 labeled examples per class, self-training with SLA and consistency regularization achieved a mean test accuracy of $94.83\%$ (std. dev. $0.32\%$) over 5 trials.
This improves on the previous state-of-the-art algorithm on this task, FixMatch, which achieved a mean test accuracy of $90.10\%$ (std. dev. $3.00\%$) with the same labeled/unlabeled splits, and is comparable to the mean test accuracy of FixMatch when trained with 25 labels per class.

The remainder of this paper is structured as follows. 
In the following section, we describe the SLA algorithm alongside a complete self-training procedure that uses SLA in its label assignment step.
After a review of related work, we present our empirical findings, which include benchmark results on the CIFAR-10, CIFAR-100, and SVHN datasets, as well as an analysis of the learning dynamics induced by SLA.
We conclude with a discussion of limitations and future work.
Our code is available at \url{https://github.com/stanford-futuredata/sinkhorn-label-allocation}.

\minihead{Notation} We denote the number of labeled examples by $n_\ell$, the total number of labeled and unlabeled examples by $n$, and the number of classes by $k$.
Let $\mathbb{R}_+$ denote the set of nonnegative real numbers, and let $\zeros{d}$ and $\ones{d}$ be the zero and all-ones vectors of dimension $d$ respectively.
Let $[m]$ denote the set of integers $\{1, \dots,  m\}$, and let $\Delta_d\coloneqq \{ x\in\mathbb{R}^d_+ \mid x^T\ones{d} = 1 \}$ denote the $d$-simplex.
Define $x_+ \coloneqq \mathrm{max}(0, x)$, and $x_- \coloneqq \mathrm{min}(0, x)$ as the positive and negative parts of~$x$.
For probability distributions $p,q\in\Delta_d$, define the entropy of $p$ as $H(p) \coloneqq -\sum_{i=1}^d p_i \log p_i$, the cross-entropy between $p$ and $q$ as $H(p, q) \coloneqq -\sum_{i=1}^d p_i \log q_i$, and the Kullback-Leibler divergence between $p$ and $q$ as $D_\mathrm{KL}(p\|q) \coloneqq \sum_{i=1}^d p_i \log(p_i / q_i)$.
We use $\langle X, Y \rangle \coloneqq \sum_{ij} X_{ij} Y_{ij}$ to denote the Frobenius inner product between matrices of equal dimension.

%% file: method.tex
\section{Sinkhorn Label Allocation (SLA)}
\label{sec:method}

We begin this section by describing the SLA optimization problem and its derivation from standard principles in semi-supervised learning.
We then show how SLA can be applied within a self-training procedure in combination with consistency regularization.

\subsection{Label Assignment}

\minihead{Soft labels}
As with any label assignment procedure, the goal of SLA is to produce a label vector $q\in\mathbb{R}^k$ for a corresponding example $x$.
SLA is a ``soft'' label assignment algorithm since it generates label vectors in the set $\{q \in \mathbb{R}^k_+ \mid q^T \ones{k} \leq 1 \}$.
While the constraint $q^T \ones{k} \leq 1$ may appear to be somewhat unusual since soft labels are typically defined to be elements of $\Delta_k$,
we note that the soft labels returned by SLA can be written as the product $q = \eta \tilde{q}$ of a distribution $\tilde{q}\in\Delta_k$ and a scalar weight $\eta\in[0,1]$.
Thus, when we plug $q$ into the standard cross-entropy loss $H(q,p)$, we obtain $H(q,p) \coloneqq -\sum_{i=1}^k q_i \log p_i = - \eta \sum_{i=1}^k \tilde{q}_i \log p_i$.
An SLA soft label therefore yields a \emph{weighted} cross-entropy loss when directly used as the target ``distribution'' during training.

\minihead{Optimization problem}
SLA derives its label assignments from the solution to the following linear program (LP):
\begin{align}
    \underset{Q \in \mathbb{R}^{n\times k}}{\mathrm{minimize}}\quad & \langle Q, C \rangle \label{eq:lp-orig} \\
    \text{s.t.}\quad& Q_{ij}\geq 0,\nonumber \\
                     & Q \ones{k}  \preceq \ones{n}, \nonumber \\ 
                     & Q^T \ones{n} \preceq \ones{k} + nb, \label{eq:col-constraint} \\ 
                     &  \ones{n}^T Q \ones{k} \geq n(\rho - \mu_+) - 1, \label{eq:mass-constraint}
\end{align}
where $C$ is the non-negative cost matrix derived from the model predictions~(Eq.~\ref{eq:cost-matrix}), $b\in\mathbb{R}^k_+$ is a vector of upper bounds on the fraction of labels that can be allocated to each class, $\rho \in [0, 1]$ is the total fraction of labels to be allocated, and $\mu \coloneqq 1 - b^T\ones{k}$. 
We subtract $\mu_+$ from $\rho$ in the mass constraint~(\ref{eq:mass-constraint}) to ensure that the problem is feasible.
We also introduce some slack to the constraints by adding 1 to each of the column constraints~(\ref{eq:col-constraint}) and subtracting 1 from the mass constraint~(\ref{eq:mass-constraint}) to ensure strict feasibility in order to avoid numerical instability in the final implementation.

We can derive the upper bound constraints from one of several sources.
Most directly, we may have prior knowledge of the label distribution, for example in settings where we have access to aggregate group-level statistics but not instance-level labels~\citep{kuck2005learning}.
Under the assumption that the labeled examples are drawn i.i.d. from the same distribution as the unlabeled examples, we may estimate upper bounds using confidence intervals for binomial proportions, e.g., the Wilson score interval~\citep{wilson1927probable}.
In settings where the unlabeled examples are sampled from a different distribution, we can estimate label proportions using methods from the domain adaptation literature~\citep{lipton2018detecting,azizzadenesheli2019regularized}.

\minihead{Derivation}
The LP formulation used in SLA (\ref{eq:lp-orig}) can be derived from standard principles in SSL.
We start by considering the following simplified label assignment problem over label distributions $Q_i \in \Delta_k$:
\begin{equation}
\label{eq:simplified-assignment}
    \underset{Q_i \in \Delta_k}{\mathrm{minimize}} \quad \sum_{i=1}^n D_\mathrm{KL}(Q_i \;\|\; P_i) + H(Q_i).
\end{equation}
This objective balances two terms: the KL-divergence term captures the requirement that the assigned labels are close to the model predictions $P_i$, while the entropy term represents the assumption that an optimal classifier should be able to unambiguously assign a class to all the unlabeled examples.
The latter implements the standard \emph{cluster assumption} that typifies many SSL algorithms, namely that the decision boundary of the classifier should only pass through low-density regions of the data distribution~\citep{joachims1999transductive,joachims2003transductive,sindhwani2006}.
The entropic penalty can also be seen to be an instance of the entropy minimization criterion in SSL~\citep{grandvaletsemi}.

Using the definition of the KL-divergence, we can rewrite the objective in (\ref{eq:simplified-assignment}) as follows:
\begin{align*}
    &\sum_{i=1}^n D_\mathrm{KL}(Q_i \;\|\; P_i) + H(Q_i) \\
    = &-\sum_{i=1}^n \sum_{j=1}^k Q_{ij} \log P_{ij} = \langle Q, C \rangle,
\end{align*}
with $C_{ij} \coloneqq -\log P_{ij}$. 
By relaxing the constraint $Q_i \in \Delta_k$ to allow partial label allocations and adding the class upper bound and total mass constraints, we obtain the LP formulation used for label assignment with SLA~(\ref{eq:lp-orig}).

\begin{algorithm}[t]
   \caption{Sinkhorn Label Allocation (SLA)}
   \label{alg:sla}
\begin{algorithmic}
    \STATE {\bfseries Input:} 
    label cost matrix~$C\in\mathbb{R}_{+}^{n\times k}$, upper bounds~$b \in \mathbb{R}^{k}_+$,
    allocation fraction~$\rho \in [0,1]$, Sinkhorn regularization parameter~$\gamma > 0$, tolerance~$\epsilon > 0$
    \STATE {\bfseries Output:} scaling variables $\alpha, \beta$
    \vspace{0.25em}
    \STATE $\alpha \gets \zeros{n+1}$, $\beta \gets \zeros{k+1}$
    \STATE $M \gets \begin{bmatrix} e^{-\gamma C} & \ones{n} \\ \ones{k}^T & 1 \end{bmatrix}$
    \vspace{0.125em}
    \STATE \textcolor{gray}{// Set target row sums $r$ and column sums $c$}
    \STATE $\mu \gets 1 - b^T \ones{k}$
    \STATE $r \gets \begin{bmatrix} \ones{n}^T & 1 + k + n(1 - \rho - \mu_-) \end{bmatrix}^T$
    \STATE $c \gets \begin{bmatrix} (\ones{k} + nb)^T & 1 + n (1 - \rho + \mu_+) \end{bmatrix}^T$
    \vspace{0.125em}
    \STATE \textcolor{gray}{// Run Sinkhorn iteration}
    \WHILE{$\| c -  M^T e^\alpha \|_1 > \epsilon$}
        \STATE $\beta \gets \log c - \log M^T e^\alpha$
        \STATE $\alpha \gets \log r - \log M e^\beta$
    \ENDWHILE
    \STATE \textbf{return} $\alpha, \beta$
\end{algorithmic}
\end{algorithm}

\minihead{Generality} This LP encodes several defining characteristics of existing label assignment procedures for self-training.
For example, suppose that we set $b = \ones{k}$ (such that constraint~(\ref{eq:col-constraint}) is vacuous), and we replace the mass constraint with $\ones{n}^T Q \ones{k} \geq n$ to ensure full allocation.
Then a solution to the LP is to set $Q_{ij} = 1$  iff $j = \arg\min_{j'} C_{ij'}$; this is the assignment scheme used in \emph{pseudo-labeling}~\citep{lee2013pseudo}.
If instead we have $\rho = 0.1$ in the mass constraint, then we have $Q_{ij} = 1$  iff $j = \arg\min_{j'} C_{ij'}$ and $x_i$ is among the $10\%$ most confidently classified examples.
The resulting allocation strategy is therefore similar to both confidence thresholding and label annealing heuristics.
Likewise, the column constraints~(\ref{eq:col-constraint}) can be used to represent class balancing heuristics frequently used in SSL.

We may additionally elect to simulate several other label assignment heuristics, e.g.:
(1) allocation upper bounds on subsets of classes instead of individual classes;
(2) time-varying column upper bounds to introduce new classes over time; and
(3) time-varying row upper bounds to simulate curriculum learning~\citep{bengio2009curriculum}, given \emph{a priori} knowledge on the difficulty of individual examples.
For simplicity, we restrict our attention in this work to the combination of label annealing and class balancing.

While the label allocation LP can be used to simulate several existing heuristics, a distinguishing property of this formulation is that it aims to optimize the label assignment globally over the entire set of unlabeled examples---this is necessary since active mass and column constraints will, in general, introduce dependencies between assignments to individual examples.

\minihead{Fast approximation} 
General-purpose LP solvers are too slow for use for label assignment within self-training due to their impractical time complexity of  $O(n^{3.5})$~\citep{renegar1988polynomial}.
Fortunately, it is possible to transform the LP in (\ref{eq:lp-orig}) to a more tractable form that is amenable to fast approximation algorithms.
We can rewrite the problem in the following equivalent form (see the Appendix for the full derivation):
\begin{align}
    \underset{Q, u, v, w}{\mathrm{minimize}} & \quad \langle Q, C \rangle  \label{eq:lp-ot}\\
    \text{s.t.} &\quad Q_{ij}\geq 0, \; u\succeq 0, \; v \succeq 0, \; \tau \geq 0, \nonumber \\
                &\quad Q \ones{k} + u = \ones{n}, \nonumber\\
                &\quad Q^T \ones{n} + v = \ones{k} + nb, \nonumber \\
                &\quad u^T \ones{n} + \tau = 1 + n(1 - \rho + \mu_+), \nonumber \\
                &\quad v^T \ones{k} + \tau = 1 + k + n(1 - \rho - \mu_-), \nonumber
\end{align}
where we have introduced additional variables $u\in\mathbb{R}^n$, $v\in\mathbb{R}^k$, and $\tau \in \mathbb{R}$.
For conciseness, we will use
\begin{equation*}
\tilde{Q} \coloneqq \begin{bmatrix} Q & u \\ v^T  & \tau \end{bmatrix}, \quad \tilde{C} \coloneqq \begin{bmatrix} C & \zeros{n} \\ \zeros{k}^T & 0 \end{bmatrix}
\end{equation*}
to denote the optimization variables and corresponding cost matrix in the problem.

By inspection, the above LP has the form of an optimal transportation problem. 
Its solution can therefore be efficiently approximated using the Sinkhorn-Knopp algorithm~\citep{cuturi2013sinkhorn,altschuler2017near}.
Given a regularization parameter $\gamma > 0$, the Sinkhorn-Knopp algorithm is an alternating projection procedure that outputs an approximate solution of the form
\begin{equation*}
    \tilde{Q} = \mathrm{diag}\left(e^\alpha\right) e^{-\gamma \tilde{C}} \mathrm{diag}\left(e^\beta\right),
\end{equation*}
where $\alpha\in\mathbb{R}^{n+1}$ and $\beta\in\mathbb{R}^{k+1}$, and exponentiation is performed elementwise.
The algorithm iteratively updates the variables $\alpha$ and $\beta$ such that the row and column marginals of $\tilde{Q}$ equal their target values.
As $\gamma \rightarrow \infty$, the solution approaches the optimum of the LP, but the alternating projection process will in turn require more iterations to converge.

Algorithm~\ref{alg:sla} summarizes the SLA label assignment process.

\subsection{Self-Training Algorithm} 

\begin{algorithm}[t]
\caption{Self-training with Sinkhorn Label Allocation and consistency regularization}
\label{alg:training-with-sla}
  \begin{algorithmic}
      \STATE {\bfseries Input:} examples~$\{x_i \mid i \in [n] \}$, labels~$\{y_i \mid i \in [n_\ell] \}$, data augmentation distributions~$P_x$, unlabeled loss weight $\lambda \geq 0$, parameter update procedure~$\textsc{ModelUpdate}$, allocation upper bounds~$b \in \mathbb{R}^{k}_+$, allocation fractions~$\rho_t\in[0, 1]$, Sinkhorn regularization parameter~$\gamma > 0$, tolerance~$\epsilon > 0$, iterations~$T$
      \STATE {\bfseries Output:} classifier $p_\theta(y\mid x)$
      \vspace{0.25em}
      \STATE Initialize model parameters $\theta_0$
      \vspace{0.125em}
      \STATE \textcolor{gray}{// Initialize scaling variables and cost matrix}
      \STATE $\beta \gets \zeros{k+1}$
      \STATE $C_{ij} \gets \log k \enspace \textbf{for}\enspace i\in[n], j\in[k]$
      \FOR{$t = 1, 2, \dots, T$}
          \STATE Sample labeled batch $\{(x_i, y_i) \mid i \in \mathcal{B}_\ell \subset [n_\ell] \}$
          \STATE Sample unlabeled batch $\{ x_i \mid i \in \mathcal{B}_u \subset [n]  \}$
          \STATE Sample augmented pairs $(\tilde{x}_i, \tilde{x}'_i)$ from $P_{x_i}$
          \vspace{0.125em}
          \STATE \textcolor{gray}{// Compute soft labels}
          \FOR{$i \in \mathcal{B}_u$}
              \STATE $p_{i} \gets p_{\theta_{t-1}} (y \mid \tilde{x}_i)$
              \STATE $q_i \gets [p_{i1}^\gamma e^{\beta_1}, \dots, p_{ik}^\gamma e^{\beta_k}, e^{\beta_{k+1}} ]$
              \STATE $q_i \gets q_i / (q_i^T \ones{k+1})$
          \ENDFOR
          \vspace{0.125em}
          \STATE \textcolor{gray}{// Compute losses and update model}
          \STATE $L_\ell(\theta) \gets -\frac{1}{|\mathcal{B}_\ell|}\sum_{i\in\mathcal{B}_\ell}\log p_\theta(y_i \mid \tilde{x}_i) $
          \STATE $L_u(\theta) \gets -\frac{1}{|\mathcal{B}_u|}\sum_{i \in \mathcal{B}_u}\sum_{j=1}^k q_{ij} \log p_\theta(j \mid \tilde{x}'_i)$
          \STATE $L(\theta) \gets L_\ell(\theta) + \lambda L_u(\theta)$
          \STATE $\theta_t \gets \textsc{ModelUpdate}(\theta_{t-1}, \nabla_\theta L)$
          \vspace{0.125em}
          \STATE \textcolor{gray}{// Update label allocation}
          \STATE $C_i \gets -\log p_i\;\textbf{for}\; i\in\mathcal{B}_u$
          \STATE $(\alpha, \beta) \gets \textsc{SLA}(C, b, \rho_t, \gamma, \epsilon)$\quad(Algorithm~\ref{alg:sla})
      \ENDFOR
      \STATE \textbf{return} $p_{\theta_T}(y\mid x)$
  \end{algorithmic}
\end{algorithm}

We can now use SLA label assignment within a self-training algorithm to instantiate a SSL procedure.
Algorithm~\ref{alg:training-with-sla} uses SLA in combination with \emph{consistency regularization}~\citep{bachman2014learning,sajjadi2016regularization,laine2016temporal}, which can be seen as a recent variant of earlier multi-view SSL approaches~\citep{blum1998combining} that penalize deviations between model predictions on perturbed instances of training examples.

In particular, Algorithm~\ref{alg:training-with-sla} incorporates the form of consistency regularization used in FixMatch~\citep{fixmatch}.
This approach samples a pair $(\tilde{x}, \tilde{x}')$ of augmented instances of an example $x$:
$\tilde{x}$ is a ``weakly augmented'' view of $x$, while $\tilde{x}'$ is a ``strongly augmented'' view corresponding to small and large perturbations of the base point respectively.
For example, a weakly augmented image may be perturbed with a small random translation, while a strongly augmented image may additionally be subject to large distortions in color.
Since we derive the soft labels $q$ solely from the weakly augmented instances $\tilde{x}$, the unlabeled loss term $L_u$ encourages predictions on the strongly augmented views to match the labels allocated to the weakly augmented views.

Algorithm~\ref{alg:training-with-sla} maintains an $n \times k$ cost matrix $C$ where each row corresponds to an unlabeled example. 
We update the entries of $C$ with the negative log probabilities assigned to each class by the current model (Eq.~\ref{eq:cost-matrix}).
To avoid incurring the computational cost of evaluating the model on the full set of examples in each iteration, we only update the rows of $C$ corresponding to the current unlabeled minibatch.

In each iteration, we derive the soft label $q$ for a given unlabeled example $x$ by rescaling the predicted label distribution using the scaling variable $\beta$ obtained from SLA:
\begin{equation}
\label{eq:rescaling}
    q_j = \frac{p_\theta(j\mid x)^\gamma e^{\beta_j}}{e^{\beta_{k+1}} + \sum_{j'=1}^{k} p_\theta(j' \mid x)^\gamma e^{\beta_{j'}}}.
\end{equation}
This rescaling is identical to that used in the Sinkhorn-Knopp algorithm.
We can interpret the additional $e^{\beta_{k+1}}$ term in the normalizer as a \emph{soft threshold}: if $e^{\beta_{k+1}} \gg p_\theta(j\mid x)^\gamma e^{\beta_j}$ for $j\in[k]$, then $q$ is close to $0$.
In such a case, we are \emph{abstaining} from assigning $x$ to a class.

The allocation schedule $\rho_t$ controls the fraction of examples that are assigned labels in each iteration.
In our experiments, we generally use a simple linear ramp from no allocation to full allocation, $\rho_t = (t-1)/(T-1)$.
In our ablation studies, we evaluate the performance of our label allocation algorithm in the absence of this ramping strategy.

%% file: related.tex
\section{Related Work}
\label{sec:related}

\minihead{Annealing and homotopy methods}
Over the course of a training run where the label allocation parameter $\rho$ is swept from 0 to 1, SLA prioritizes the highest-confidence predictions in its label assignments.
This assignment strategy is reminiscent of curriculum learning~\citep{bengio2009curriculum} and self-paced learning~\citep{kumar2010self}, where ``easy'' examples are used early in training and more ``difficult'' examples are gradually introduced over time.
As with these other methods, self-training with SLA can be interpreted as a \emph{homotopy} or \emph{continuation method} for nonconvex optimization~\citep{allgower1990numerical}, which iteratively solve a sequence of relaxed problem instances that eventually converges to the original optimization problem. 
In the context of SSL, \citet{sindhwani2006} propose a homotopy strategy for training semi-supervised SVMs that gradually anneals the entropy of soft labels assigned to the unlabeled examples---this strategy differs from our approach since it involves an assignment of labels to \emph{all} unlabeled examples in each iteration.

The confidence thresholding heuristic used in FixMatch~\citep{fixmatch} also induces an annealing schedule: as model predictions become more confident over the course of training, unlabeled examples are more frequently assigned labels and thus more frequently contribute to model updates.\footnote{We document this effect empirically in Sec.~\ref{sec:dynamics}.}
However, it is generally unclear how the confidence threshold should be set since the predictions of many modern neural network architectures are known to not be calibrated without additional post-processing~\citep{hendrycks2016baseline,guo2017calibration}.
Our use of an allocation schedule in SLA obviates the need to manually select a confidence threshold parameter for training.

\minihead{Robust estimation}
The bootstrapping process in self-training is essentially a problem of learning with noisy labels where the source of label noise is the inaccuracy of the classifier during training, in contrast to the typical assumptions of random or adversarial label corruption. 
We can view the label annealing component of SLA as a means of mitigating label noise---from this perspective, the SLA label assignment process is similar to robust learning methods such as iterative trimmed loss minimization~\citep{shen2019learning}, which computes model updates using only a preset fraction of low-loss training examples.

\minihead{Class balancing}
The use of class balancing criteria has long been commonplace in SSL algorithms in order to avoid imbalanced label assignments.
The original co-training algorithm~\citep{blum1998combining} grows the training set by adding artificially labeled examples in proportion to the class ratio in the labeled set, while the Transductive SVM~\citep{joachims1999transductive} fixes the number of positive labels to be assigned to the unlabeled data.
Variants of class balancing have since appeared in many other works~\citep{zhu2002,sindhwani2006,chapelle2008optimization}.
A recent example is the ReMixMatch algorithm, which employs a variant of class balancing called ``distribution alignment''~\citep{kurakin2020remixmatch}.
In self-supervised learning, Sinkhorn iteration has been used to ensure an even assignment of examples to clusters~\citep{asano2020self,caron2020unsupervised}.
A distinguishing feature of SLA is its use of upper bounds instead of exact equality constraints, which allows for additional flexibility in the label assignment process.

Our class proportion constraints are also similar to prior work on learning from label proportions~\citep{kuck2005learning,musicant2007supervised,dulac2019}, where the goal is to learn a classifier given the label distributions of several subsets of examples.
Our setting involves a single global set of constraints on the class distribution of the unlabeled set, in contrast to the LLP setting which concerns large sets of small bags of data.

Additionally, class proportion constraints are also conceptually related to methods for learning with constraints on the model posterior, e.g., constraint driven learning~\citep{chang2007guiding}, generalized expectation criteria~\citep{mann2007simple,mann2008generalized}, and posterior regularization~\citep{ganchev2010posterior}.
These methods aim to guide learning by constraining posterior expectations of user-defined features that encode prior knowledge about the desired solution.

\minihead{Expectation Maximization}
Finally, we remark that the alternating minimization process in Algorithm~\ref{alg:training-with-sla} that iterates between label updates and model updates is similar to applications of the EM algorithm in SSL~\citep{nigam2000text}.
Our algorithmic approach differs since we do not use label expectations with respect to a probabilistic model.

%% file: experiments.tex
\section{Experiments}

\begin{table*}[t]
\caption{A test error comparison (mean and standard deviation over 5 runs) on CIFAR-10 and CIFAR-100 with varying labeled set sizes.
We obtained the FixMatch results using our own reimplementation, while the results for MixMatch~\citep{berthelot2019mixmatch}, UDA~\citep{xie2019unsupervised}, and ReMixMatch~\citep{kurakin2020remixmatch} are as reported in~\citep{fixmatch}.
SLA improves on the mean accuracy of FixMatch on CIFAR-10 and CIFAR-100 for all labeled set sizes, except for the 2500 label runs on CIFAR-100.
}
\vspace{-1em}
\label{tab:c10-c100-comparison}
\begin{center}
\resizebox{\textwidth}{!}{
\sisetup{table-format=2.2(1)}
\begin{tabular}{lSSSSSSSS}
\toprule
&&& \multicolumn{1}{c}{\textbf{CIFAR-10}} &&& \multicolumn{3}{c}{\textbf{CIFAR-100}} \\
  \cmidrule(lr){2-6} \cmidrule(lr){7-9}
\textbf{Method} & \multicolumn{1}{c}{10 labels} & \multicolumn{1}{c}{20 labels} & \multicolumn{1}{c}{40 labels} & \multicolumn{1}{c}{80 labels} & \multicolumn{1}{c}{250 labels} & \multicolumn{1}{c}{400 labels} & \multicolumn{1}{c}{800 labels} & \multicolumn{1}{c}{2500 labels} \\ 
\cmidrule(lr){1-1} \cmidrule(lr){2-6} \cmidrule(lr){7-9}
MixMatch & \multicolumn{1}{c}{-} & \multicolumn{1}{c}{-} & 47.54\pm11.50 & \multicolumn{1}{c}{-} & 11.05\pm0.86 & 67.61\pm1.32 & \multicolumn{1}{c}{-} & 39.94\pm0.37 \\
UDA & \multicolumn{1}{c}{-} & \multicolumn{1}{c}{-} & 29.05\pm5.93 & \multicolumn{1}{c}{-} & 8.82\pm1.08 & 59.28\pm0.88 & \multicolumn{1}{c}{-} & 33.13\pm0.22 \\
ReMixMatch & \multicolumn{1}{c}{-} & \multicolumn{1}{c}{-} & 19.10\pm9.64 & \multicolumn{1}{c}{-} & 5.44\pm0.05 & 44.28\pm2.06 & \multicolumn{1}{c}{-} & 27.43\pm0.31 \\
\midrule
FixMatch & 37.02\pm 8.35 & 20.53\pm8.90 & 9.90\pm3.00 & 6.42\pm0.21 & 5.09\pm 0.61 & 43.42\pm2.41 & 35.53\pm1.00 & 27.99\pm0.42 \\ 
SLA      & 34.13\pm10.83 & 18.09\pm6.77 & 5.17\pm0.32 & 5.02\pm0.28 & 4.89\pm 0.27 & 41.44\pm1.41 & 34.31\pm1.09 & 28.73\pm0.44 \\ 
\bottomrule
\end{tabular}
}
\end{center}
\vspace{-1em}
\end{table*}

In this empirical study, we investigate (1) the accuracy of classifiers trained with SLA, (2) the training dynamics induced by the SLA label assignment process, and (3) the effect the hyperparameters introduced by SLA.
Our main baseline for comparison is the FixMatch algorithm~\citep{fixmatch} since it is a state-of-the-art method for semi-supervised image classification.
For each configuration, we report the mean and standard deviation of the error rate across 5 independent trials.

\minihead{Datasets and labeled splits}
We used the CIFAR-10, CIFAR-100~\citep{krizhevsky2009learning}, and SVHN~\citep{netzer2011reading} image classification datasets with their standard train/test splits.
In each trial, we independently sampled a labeled set without replacement from the training split, and we used the same labeled/unlabeled splits across runs of different methods.
We used labeled set sizes of  $\{10, 20, 40, 80, 250\}$ for CIFAR-10, $\{400, 800, 2500\}$ for CIFAR-100, and $\{20, 40, 80\}$ for SVHN.

Following the experimental protocol in recent work~\citep{kurakin2020remixmatch,fixmatch}, we chose the label distribution of the labeled set such that it is as close as possible to the true label distribution of the training set in total variation distance, subject to the constraint that there is at least one example sampled for each class.
We observe that this setup implies that the empirical label distributions of the labeled sets for CIFAR-10/100 are always \emph{well-specified}, in the sense that they are equal to the true distribution of labels in the training set.\footnote{This is due to our choices of labeled set sizes, and that CIFAR-10/100 are balanced datasets.}
In contrast, the empirical label distributions for SVHN are \emph{misspecified} since the training label distribution is non-uniform.\footnote{The TV distances for SVHN with 20, 40, and 80 labels are $0.068$, $0.034$, and $0.018$ respectively.}
Since the well-specified setting is arguably somewhat unrealistic for real-world SSL applications, we additionally report the results of CIFAR-10 experiments in the misspecified case where the labeled sets are sampled uniformly without replacement from the training split, conditioned on there being at least one example per class.

\minihead{Hyperparameters}
Our experiments used the same experimental setup as in the evaluation of FixMatch where applicable.
We optimized our classifiers using the stochastic Nesterov accelerated gradient method with a momentum parameter of $0.9$ and a cosine learning rate schedule given by $0.03 \cos(7\pi t / 16 T)$, where $t$ is the current iteration and $T=2^{20}$ is the total number of iterations.\footnote{This schedule anneals the learning rate from $0.03$ to $\approx 0.006$.}
We used a labeled batch size of $64$, an unlabeled batch size of $448$, weight decay of $5\times 10^{-4}$ on all parameters except biases and batch normalization weights, and unlabeled loss weight $\lambda = 1$.
For CIFAR-10 and SVHN, we used the Wide ResNet-28-2 architecture~\citep{Zagoruyko2016WRN}, whereas for CIFAR-100, we used the Wide ResNet-28-8 architecture (with a weight decay of $10^{-3}$).
When evaluating on the test set, we used an exponential moving average of the model parameters~\citep{tarvainen2017mean} with a decay parameter of $0.999$.
We used a confidence threshold of $0.95$ for our FixMatch baselines.

For hyperparameters specific to SLA, we used an Sinkhorn regularization parameter of $\gamma = 100$ and tolerance parameter $\epsilon_t = 0.01 \|c_t\|_1$ for Sinkhorn iteration, where $c_t$ is the target column sum at iteration $t$.
Unless otherwise specified, we increased the allocation parameter $\rho$ linearly from $0$ to $1$ over the course of training.
For CIFAR-10/100, we used the empirical label distribution of the labeled examples as the class proportion upper bounds $b$. 
For SVHN, we used upper bounds given by the $80\%$ Wilson score interval~\citep{wilson1927probable} since the empirical label distribution only approximates the true label distribution.

\minihead{Data augmentation}
We ran both SLA self-training and the FixMatch baselines with the same data augmentation distributions.
For consistency regularization, our weak augmentation policy consisted of random translations of up to 4 pixels (for all datasets) and random horizontal flips with probability $0.5$ (for CIFAR-10/100, but not SVHN).
Our strong augmentation policy consisted of the weak augmentation policy composed with RandAugment~\citep{cubuk2020randaugment}, followed by $16\times16$ Cutout augmentations~\citep{devries2017improved}.

\minihead{Computational cost}
In our runs, SLA incurred an average $21.1\%$ overhead in total training time for CIFAR-10 and a $23.2\%$ overhead for CIFAR-100.

\subsection{Classification Benchmarks}

\begin{table}
\caption{A test error comparison on SVHN with varying labeled set sizes.
The results for MixMatch, UDA, and ReMixMatch are as reported in~\citep{fixmatch}.
SLA improves on FixMatch on average, except with 20 labeled examples where the class upper bounds are poor estimates of the true label distribution. 
}
\label{tab:svhn-comparison}
\begin{center}
\resizebox{0.46\textwidth}{!}{
\sisetup{table-format=2.2(1)}
\begin{tabular}{lSSS}
\toprule
&& \multicolumn{1}{c}{\textbf{SVHN}} \\
\cmidrule(lr){2-4}
\textbf{Method} & \multicolumn{1}{c}{20 labels} & \multicolumn{1}{c}{40 labels} & \multicolumn{1}{c}{80 labels} \\ 
\cmidrule(lr){1-1} \cmidrule(lr){2-4}
MixMatch & \multicolumn{1}{c}{-} & 42.55\pm14.53 & \multicolumn{1}{c}{-} \\
UDA & \multicolumn{1}{c}{-} & 52.63\pm20.51 & \multicolumn{1}{c}{-} \\
ReMixMatch & \multicolumn{1}{c}{-} & 3.34\pm0.20 & \multicolumn{1}{c}{-} \\
\midrule
FixMatch & 14.92\pm7.82 & 4.74\pm3.28 & 2.98\pm1.31 \\
SLA      & 22.85\pm9.84 & 3.63\pm2.91 & 2.48\pm0.18 \\
\bottomrule
\end{tabular}
}
\end{center}
\vspace{-1.5em}
\end{table}

\begin{table}
\caption{A test error comparison on CIFAR-10 with 40 labels distributed evenly between the classes (Uniform) and with 40 labels sampled uniformly from the training set, conditioned on at least one label being drawn for each class (Multinomial).
Accuracy degrades for all methods in the more challenging multinomial setting.
}
\label{tab:multinomial-comparison}
\begin{center}
\resizebox{0.48\textwidth}{!}{
\sisetup{table-format=2.2(1)}
\begin{tabular}{lSS}
\toprule
\textbf{Method} & \multicolumn{1}{c}{Uniform} & \multicolumn{1}{c}{Multinomial} \\ 
\cmidrule(lr){1-1} \cmidrule(lr){2-3}
FixMatch & 9.90\pm 3.00     & 11.23\pm3.56 \\
FixMatch (with DA) & 5.70\pm1.63  & 18.64\pm11.29 \\
\midrule
SLA (without upper bounds)      & 9.71\pm5.95  & 13.40\pm6.41 \\
SLA   & 5.17\pm0.32  & 14.95\pm7.12 \\
\bottomrule
\end{tabular}
}
\end{center}
\vspace{-1.5em}
\end{table}

Tables~\ref{tab:c10-c100-comparison} and \ref{tab:svhn-comparison} summarize the test error rates achieved by self-training with FixMatch and SLA on CIFAR-10, CIFAR-100 and SVHN.
We observe an improvement in mean accuracy over FixMatch on the CIFAR-10 dataset across all configurations, on CIFAR-100 with 400 and 800 labels, and on SVHN with 40 and 80 labels.
In particular, the accuracy of SLA on CIFAR-10 with 40 labels ($94.83\%$) was comparable to the accuracy of FixMatch on 250 labels ($94.91\%$).

SLA often yielded more consistent results across runs; for example, the standard deviation for CIFAR-10 with 40 labels was reduced by $2.7\%$, and for SVHN with 80 labels by $1.1\%$. 
This can be attributed to the use of the upper bound constraints, which help prevent convergence to poor local minima due to the overrepresentation of certain classes during training.

Table~\ref{tab:multinomial-comparison} compares test errors on CIFAR-10 with 40 labels, where the empirical label distribution of the labeled set is well-specified (Uniform) or misspecified (Multinomial).\footnote{The mean TV distance to the true label distribution in the multinomial setting is $\approx 0.154$.}
We compare SLA with and without the class proportion upper bounds against standard FixMatch and FixMatch with the distribution alignment (DA) heuristic~\citep{kurakin2020remixmatch} that encourages the model label distribution to match the empirical label distribution.
In the multinomial setting, we used $80\%$ Wilson upper bounds for SLA.
As expected, the performance of all four methods degrades in the more challenging multinomial setting.
FixMatch with DA incurs a large misspecification penalty since DA essentially imposes a soft equality constraint with the empirical label distribution. 
In comparison, SLA incurs a smaller accuracy penalty due to its more forgiving upper bound constraints.

\subsection{Training Dynamics}
\label{sec:dynamics}

\begin{figure*}[t]
    \centering
    \includegraphics[width=\textwidth]{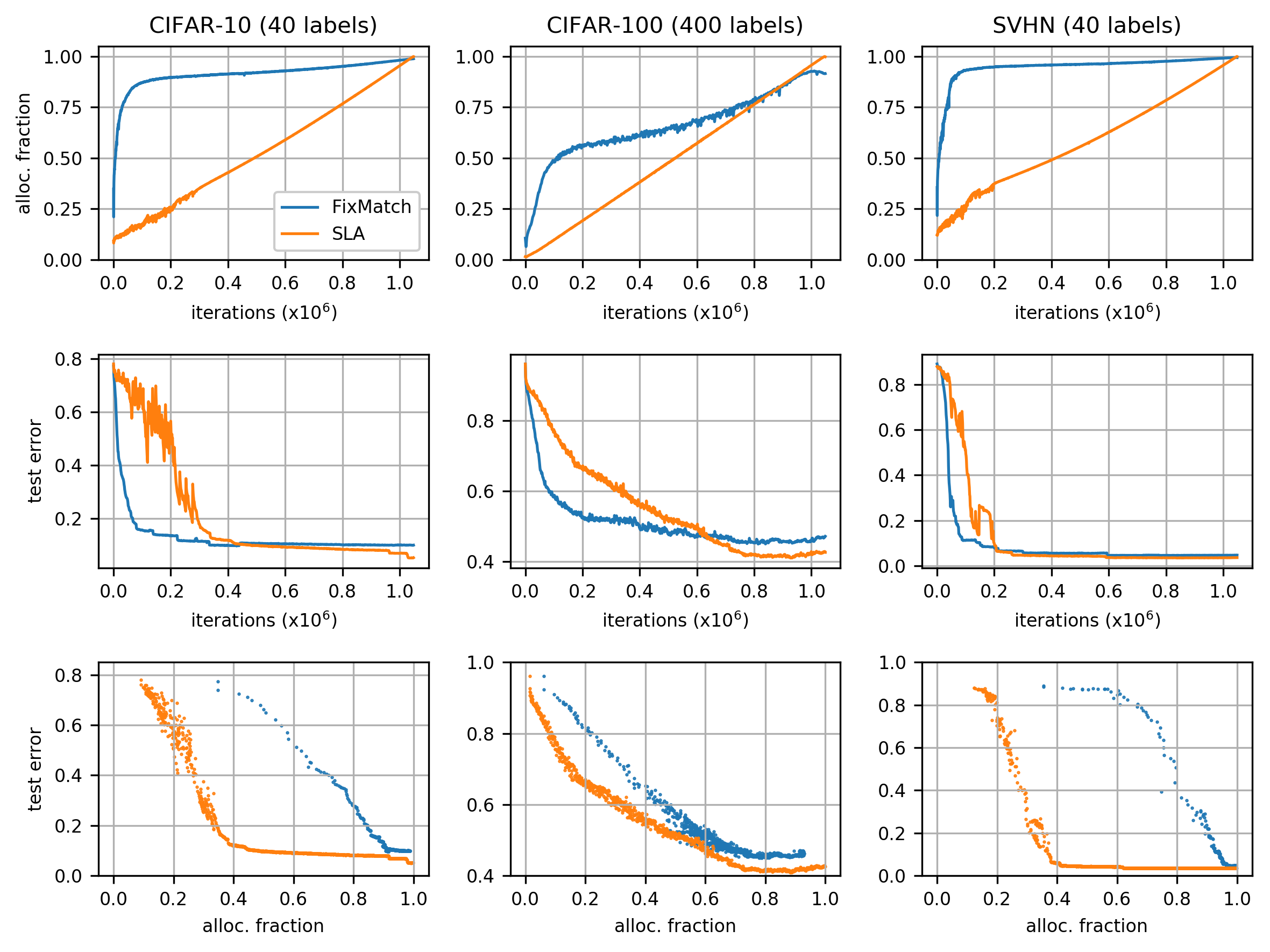}
    \vspace{-2.5em}
    \caption{The fraction of unlabeled examples assigned labels over the course of training (top row), test error during training (middle row), and the relationship between the allocation fraction and the test error during training (bottom row).
    FixMatch induces an annealing schedule that quickly increases the allocation fraction early in training, while SLA allocation increases approximately linearly according to the $\rho_t$ schedule (the SLA allocation is not exactly linear since the mass constraint is a lower bound).
    In these experiments, SLA yields lower test error on average across all allocation fractions.}
    \label{fig:err-alloc}
    \vspace{-0.5em}
\end{figure*}

Figure~\ref{fig:err-alloc} shows the total fraction of unlabeled examples that are assigned labels as a function of the training iteration count.
These plots show that the FixMatch confidence thresholding criterion induces an implicit annealing schedule where the allocated fraction increases quickly early in training.
In fact, FixMatch never reaches full label allocation with its fixed confidence threshold in the case of CIFAR-100 with 400 labels.
We suggest that the explicit allocation schedule used in SLA is a more intuitive interface for practitioners than the fixed confidence threshold used in FixMatch.

In the bottom row of Figure~\ref{fig:err-alloc}, we observe that SLA typically achieves higher test accuracy at any fixed allocation fraction.
Further, we note that the effect of the SLA constraints is apparent in the CIFAR-10 runs, where the accuracy improves in a stepwise fashion towards the end of training as the remaining ``difficult'' examples are assigned labels.

For CIFAR-100 (middle column), we find that the test error for SLA reaches a minimum and then increases towards the end of training.
This ``U''-shaped test error rate suggests that in some settings, label noise due to misclassification can start to dominate as we approach full allocation.
This observation indicates that partial label allocation, e.g. with a truncated schedule such as $\rho_t = \min\left(0.8, \frac{t-1}{T-1}\right)$, can be an effective strategy for certain tasks.

\begin{figure}[t]
    \centering
    \includegraphics[width=0.5\textwidth]{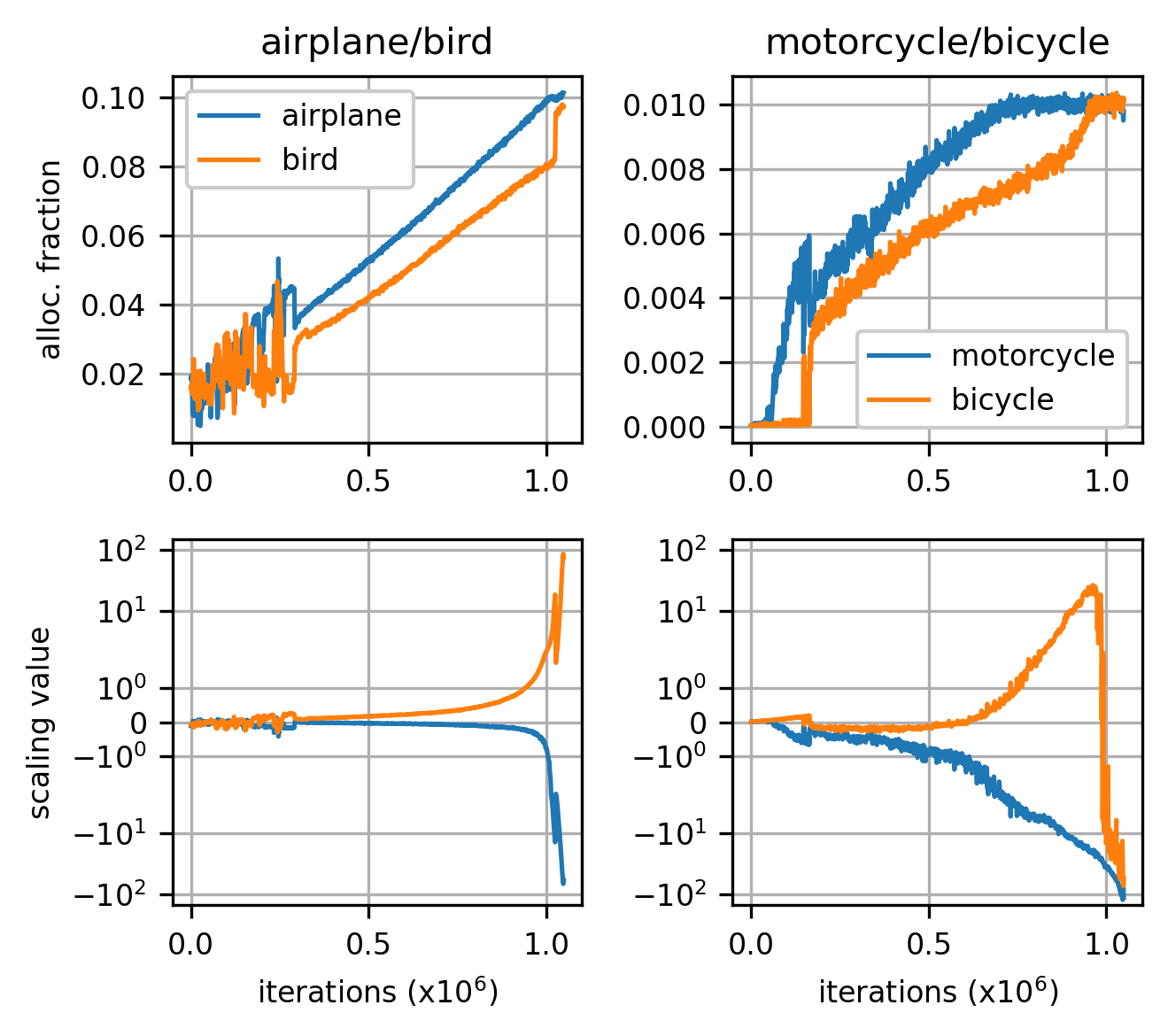}
    \vspace{-2.5em}
    \caption{SLA scaling variables $\beta$ and class allocations over the course of training for two similar class pairs: airplane/bird (CIFAR-10) and motorcycle/bicycle (CIFAR-100).
    The scaling variables promote underrepresented classes (positive values) and inhibit overrepresented classes (negative values).
    The \emph{bicycle} class is initially promoted, but as it approaches the allocation constraint of $0.01$, the corresponding scaling variable turns negative in order to enforce the upper bound.
    }
    \label{fig:scaling-vars}
    \vspace{-1em}
\end{figure}

In Figure~\ref{fig:scaling-vars}, we plot the values of the scaling values $\beta$ and the label allocations corresponding to two pairs of similar classes from CIFAR-10 and CIFAR-100. 
These plots illustrate the role of the scaling variables in influencing the dynamics of training by promoting underrepresented classes and inhibiting overrepresented classes. 
Indeed, this is consistent with their interpretation as dual variables corresponding to the class balancing constraints in the optimization problem.

\subsection{Ablations}

We investigate the effects of SLA-specific hyperparameters through a series of ablation experiments.
First, the use of a label annealing strategy is important: without any label annealing, i.e., by setting $\rho_t = 1$, we achieve a test error of $13.67 \pm 1.83\%$ on CIFAR-10 with 40 labels (vs. $5.17\pm0.32\%$ with the default linear ramp).
The use of class proportion upper bounds has a significant positive effect when the label distribution is well-specified: removing these class constraints while retaining label annealing achieves $9.71\pm 5.95\%$.

Larger values of the Sinkhorn regularization parameter $\gamma$ result in a better approximation to the solution of the optimal transportation problem, at the cost of additional time spent on Sinkhorn iteration. 
We find that the use of an overly coarse approximation has a significant negative effect on final accuracy.
Specifically, $\gamma = 1, 10, 100, 1000$ achieve single-run error rates of $42.48$, $5.78$, $4.94$, and $5.10\%$ respectively on CIFAR-10 with 40 labels.
For our set of tasks, we find that $\gamma = 100$ strikes an acceptable trade-off between approximation accuracy and speed.

%% file: discussion.tex
\section{Discussion}

In this work, we motivated SLA as an optimization-based strategy for assigning labels in self-training.
This framework proved to be sufficiently rich to synthesize several existing label assignment heuristics in SSL under a single formulation, while still retaining computational tractability via the use of an efficient approximation algorithm.

An attractive direction for future work is to extend the flexibility of this general optimization framework by allowing for a wider range of constraints, thus allowing for the incorporation of richer forms of prior knowledge in SSL problems.
A possible extension of SLA is to replace the Sinkhorn-Knopp iteration with Dykstra's algorithm~\citep{dykstra1985iterative,benamou2015iterative}, which performs cyclic Bregman projections onto collections of convex sets.
An example usecase that would be enabled by such an extension would be \emph{semi-supervised multi-label learning}.
This setting corresponds to a simple modification of our LP constraints: we stipulate $0 \leq Q_{ij} \leq 1$ and replace the constraint $Q\ones{k} \preceq \ones{n}$ with $Q\ones{k} \preceq N\ones{n}$, where $N$ is the maximum number of labels that can be assigned to an example.
Another possible usecase is the introduction of \emph{lower bounds} on class proportions in addition to our current upper bound constraints.
The use of lower bounds would help prevent one of the failure modes we observed in our experiments, namely where a subset of classes end up with zero allocation when using loose upper bounds or partial label allocation.
These lower bound constraints are not handled by our current reduction to optimal transport.

In our experiments, we found that SLA is susceptible to incurring an accuracy penalty when the constraints are misspecified.
In a sense, this should not be surprising as it can be understood as a manifestation of the ``no free lunch'' theorems.
However, we nevertheless speculate that it may be possible to extend SLA such that it is able to adaptively identify possible misspecification over the course of training.
For instance, we observe empirically that infeasible or near-infeasible constraints result in a chaotic oscillation of the model parameters and scaling variables---such signals may potentially be used to dynamically tune the constraint set during training.
Orthogonally, we note that methodological advancements on the problem of estimating label proportions from unlabeled data can yield immediate improvements for SLA via tighter bounds on class proportions.

%% file: appendix.tex
\begin{appendix}
\section{Derivation of the Optimal Transport LP}
\label{sec:lp-derivation}

We begin with the original assignment LP~(\ref{eq:lp-orig}):
\begin{align*}
    \underset{Q}{\mathrm{minimize}}\quad&\langle Q, C \rangle \\
    \text{s.t.}\quad&  Q_{ij}\geq 0,\\ 
    & Q \ones{k} \preceq \ones{n},\\ 
    & Q^T \ones{n} \preceq \ones{k} + nb,\\ 
    &\ones{n}^T Q \ones{k}\geq n(\rho - \mu_+) - 1,
\end{align*}
where $\mu \coloneqq 1 - b^T\ones{k}$. We can replace the inequality constraints on the marginals and the total assigned mass by introducing non-negative slack variables $u$, $v$, and $\tau$. This yields the following equivalent optimization problem:
\begin{align}
    \underset{Q,u,v,\tau}{\mathrm{minimize}}\quad& \langle Q, C \rangle \nonumber \\
    \text{s.t.}\quad&  Q_{ij}\geq 0, u \succeq 0, v \succeq 0, \tau \geq 0, \nonumber \\
    & Q \ones{k} + u = \ones{n}, \label{eq:cons1}\\ 
    & Q^T \ones{n} + v = \ones{k} + nb, \label{eq:cons2}\\ 
    &\ones{n}^T Q \ones{k} = \tau + n(\rho - \mu_+) - 1 \label{eq:cons3}.
\end{align}
We now rewrite the constraints to eliminate the total mass term. Substituting (\ref{eq:cons1}) into (\ref{eq:cons3}), we obtain:
\begin{align*}
    \ones{n}^T u + \tau = 1 + n(1 - \rho + \mu_+).
\end{align*}
Substituting (\ref{eq:cons2}) into (\ref{eq:cons3}), we obtain:
\begin{align*}
    \ones{k}^T v + \tau &= 1 + k + n(\ones{k}^T b - \rho)\\ &= 1 + k + n(1 - \rho - \mu_-).
\end{align*}
Thus, (\ref{eq:lp-orig}) is equivalent to the following LP:
\begin{align*}
    \underset{Q, u, v, \tau}{\mathrm{minimize}} & \quad\langle Q, C \rangle \\
    \text{s.t.} &\quad Q_{ij}\geq 0, \; u\succeq 0, \; v \succeq 0, \; \tau \geq 0, \\
                &\quad Q \ones{k} + u = \ones{n},\\
                &\quad Q^T \ones{n} + v = \ones{k} + nb, \\
                &\quad u^T \ones{n} + \tau = 1 + n(1 - \rho + \mu_+),\\
                &\quad v^T \ones{k} + \tau = 1 + k + n(1 - \rho - \mu_-),
\end{align*}
which we recognize as an optimal transportation problem with marginals $r \coloneqq \begin{bmatrix} \ones{n}^T & 1 + k + n(1 - \rho - \mu_-) \end{bmatrix}^T$ and $c \coloneqq \begin{bmatrix} \ones{k}^T & 1 + n (1 - \rho + \mu_+) \end{bmatrix}^T$.

\end{appendix}